\documentclass {elsarticle}
\pdfoutput=1
\usepackage{listings}
\usepackage{url}
\usepackage[linesnumbered,ruled,vlined]{algorithm2e}
\usepackage{float}
\usepackage{graphicx}
\usepackage[T1]{fontenc}	
\usepackage{amsmath}		
\usepackage{amssymb}		
\usepackage{multirow}
\restylefloat{figure}

\newcommand{\R}{{\mathbb R}}

\newcommand{\be}{\begin{equation}}
\newcommand{\ee}{\end{equation}}
\newcommand{\ba}{\begin{array}}
\newcommand{\ea}{\end{array}}
\newcommand{\baa}{\left[\begin{array}}
\newcommand{\eaa}{\end{array}\right]}

\newcommand{\beqa}{\begin{eqnarray}}
\newcommand{\eeqa}{\end{eqnarray}}
\newcommand{\bt}{\begin{tabular}}
\newcommand{\et}{\end{tabular}}

\newcommand{\bi}{\begin{itemize}}

\newcommand{\ei}{\end{itemize}}
\newcommand{\bc}{\begin{center}}
\newcommand{\ec}{\end{center}}

\begin {document}
\begin{frontmatter}

\title{Overcomplete Dictionary Learning with Jacobi Atom Updates}

\author{Paul~Irofti}
\ead{paul@irofti.net}

\author{Bogdan~Dumitrescu}
\ead{bogdan.dumitrescu@acse.pub.ro}

\tnotetext[thanks]{
	This work was supported by the Romanian National Authority
	for Scientific Research, CNCS - UEFISCDI,
	project number PN-II-ID-PCE-2011-3-0400,
	and by the
	Sectoral Operational Programme Human Resources Development 2007-2013
	of the Ministry of European Funds through the
	Financial Agreement POSDRU/159/1.5/S/132395.
	\newline \indent
	B.~Dumitrescu is also with Department of Signal Processing,
	Tampere University of Technology, Finland.
}
\address{
	Department of Automatic Control and Computers \\
	University Politehnica of Bucharest \\
	313 Spl. Independen\c{t}ei, 060042 Bucharest, Romania
}

\begin{abstract}
Dictionary learning for sparse representations is traditionally approached
with sequential atom updates, in which an optimized atom is used immediately
for the optimization of the next atoms.
We propose instead a Jacobi version, in which groups of atoms are updated
independently, in parallel.
Extensive numerical evidence for sparse image representation
shows that the parallel algorithms, especially when all atoms are updated
simultaneously, give better dictionaries than their
sequential counterparts.
\end{abstract}


\end{frontmatter}
\section{Introduction}

The sparse representations field is the basis for a
wide range of very effective signal processing techniques with
numerous applications for, but not limited to, audio and image
processing.

In this article, we approach the problem of training dictionaries for sparse
representations by learning from a representative data set.
Given a set of signals $Y \in \R^{p \times m}$ and a sparsity level $s$,
the goal is to find a dictionary $D \in \R^{p \times n}$
that minimizes the Frobenius norm of the approximation error
\be
E = Y - DX,
\label{eq:frob}
\ee
where $X \in \R^{n \times m}$ is the associated $s$-sparse representations matrix,
with at most $s$ nonzero elements on each column.
Otherwise said, each column (signal or data vector) from $Y$ is represented as
a linear combination of at most $s$ columns (atoms) from $D$.
To eliminate the magnitude ambiguity in this bilinear problem, where both $D$ and $X$
are unknown, the columns of the dictionary are constrained to unit norm.

Since dictionary learning (DL) for sparse representations is a hard problem,
the most successful algorithms, like K-SVD \cite{AEB06} (and its approximate
version AK-SVD \cite{RZE08}) and
MOD \cite{EAH99mod}, adopt an alternating optimization procedure with two basic stages.
First, fixing the current dictionary $D$ (initialized randomly or with a subset of $Y$),
the sparse representations $X$ are computed with Orthogonal Matching Pursuit (OMP) \cite{PRK93omp}
or another algorithm.
Then, keeping $X$ fixed, a new dictionary is obtained through various techniques.
The second stage, where the atoms of the dictionary are updated, makes the
main difference between DL algorithms.
Recent methods or improvements can be found in \cite{SmEl13}, \cite{SM13}, \cite{SBJ13}, \cite{SBJ14}.
Some of them will be discussed later, since they are used for supporting our method.
Among the algorithms related to DL, but with more constraints on the dictionary,
are \cite{RuDu12spl}, \cite{SDB12}, \cite{BaPl13}.
Overviews of earlier work and applications are presented in \cite{RBE10,ToFr11}.

All these DL algorithms update the atoms one by one, in Gauss-Seidel style.
The motivation is the classical one: an updated atom, assumed to be better
than its previous value, can be used immediately for other updates.
We investigate here the Jacobi version of several algorithms, where groups of
atoms are updated simultaneously.
We started this work in \cite{IrDu14}, where our study was confined to AK-SVD,
aiming at reducing the dictionary design time on a GPU architecture.
However, extensive numerical evidence shows that not only this strategy is not
worse than the standard sequential approach, but in many circumstances gives a smaller
representation error (\ref{eq:frob}).
This manuscript presents the Jacobi atom updates (JAU) strategy in section \ref{sec:jau},
its particular form for a few of the best sequential methods in section \ref{sec:palg}
and the above mentioned numerical evidence in section \ref{sec:res}.

As a side remark, the name "parallel atom updates" (PAU) is at least as good as JAU
to label our approach. Unfortunately, this name was already used in \cite{SBJ14}
although the atoms are updated sequentially there, using several AK-SVD update sweeps.
An idea similar with PAU is called more appropriately "dictionary update cycles" in \cite{SmEl13},
in the context of K-SVD.

\section{Jacobi Atom Updates Stategy}
\label{sec:jau}

The general form of the proposed dictionary learning method
with Jacobi atom updates is presented in Algorithm~\ref{algo:dl-jau}.
At iteration $k$ of the DL method, the two usual stages are performed.
In step 1, the current dictionary $D^{(k)}$ and the signals $Y$ are
used to find the sparse representation matrix $X^{(k)}$ with $s$ nonzero
elements on each column; we used OMP, as widely done in the literature.

The atom update stage takes place in groups of $\tilde n$ atoms.
We assume that $\tilde n$ divides $n$ only for the simplicity of description,
but this is not a mandatory condition.
Steps 2 and 3 of Algorithm~\ref{algo:dl-jau} perform a full sweep of the atoms.
All the $\tilde n$ atoms from the same group are updated independently (step 4),
using one of the various available rules; some of them will be discussed in the
next section.
Once a group is processed, its updated atoms are used for updating the other atoms;
so, atom $d_j^{(k+1)}$ (column $j$ of $D^{(k+1)}$) is computed in step 4 using $d_i^{(k+1)}$ if
\be
\lfloor (i-1) / \tilde n \rfloor < \lfloor (j-1) / \tilde n \rfloor,
\label{ijupdate}
\ee
i.e.\ $i < j$ and $d_i$ not in the same group as $d_j$, and $d_i^{(k)}$ otherwise.

Putting $\tilde n = 1$ gives the usual sequential Gauss-Seidel form.
Taking $\tilde n = n$ leads to a fully parallel update, i.e.\ the form that
is typically labeled with Jacobi's name.

The specific atom update strategy of each algorithm is contained in step 4
while step 5 is the usual normalization constraint on the dictionary.

The proposed form has obvious potential for a smaller execution time
on a parallel architecture.
We lightly touch this issue here and provide comparative execution times
from a few experiments in section \ref{sec:res};
the reader can consult \cite{IrDu14}
for an in-depth analysis of the GPU implementation of the AK-SVD algorithm.
Our main focus here is on the quality of the designed dictionary.

\begin{algorithm}[t]
\label{algo:dl-jau}
\SetKwComment{Comment}{}{}

\BlankLine
\KwData{\quad current dictionary $D^{(k)} \in \R^{p \times n}$ \\
	\hspace{13.6mm} signals set $Y \in \R^{p \times m}$ \\
	\hspace{13.6mm} number of parallel atoms $\tilde n$}
\KwResult{\hspace{1mm}next dictionary $D^{(k+1)}$}
\BlankLine

Compute $s$-sparse representations $X^{(k)} \in \R^{n \times m}$ such
that $Y \approx D^{(k)} X^{(k)}$ \\
\For{$\ell = 1$ \KwTo $n / \tilde n$}{
	\For{$j = (\ell-1) \tilde n + 1$ \KwTo $\ell \tilde n$, in parallel}{
	Update $d_j^{(k+1)}$ \\
	Normalize: $d_j^{(k+1)} \leftarrow d_j^{(k+1)} / \| d_j^{(k+1)} \|$
		}
	}
\caption{General structure of a DL-JAU iteration}
\end{algorithm}

\section{Particular forms of the algorithm}
\label{sec:palg}

Typically, the atom update problem is posed as follows.
We have the dictionary, denoted generically $D$, and the associated representations
matrix $X$ and we want to optimize atom $d_j$.
In the DL context, at iteration $k$ of the learning process,
the dictionary is made of atoms from $D^{(k)}$ and $D^{(k+1)}$, as explained
by the phrase around equation (\ref{ijupdate}).
We denote ${\cal I}_j$ the (column) indices of the signals that use $d_j$
in their representation, i.e.\ the indices of the nonzero elements on the $j$-th row of $X$.
Excluding atom $d_j$, the representation error matrix (\ref{eq:frob}), reduced
to the relevant columns, becomes
\be
F = E_{{\cal I}_j} + d_j x_{j,{\cal I}_j}.
\label{Ferror}
\ee
The updated atom $d_j$ is the solution of the optimization problem
\be
\begin{aligned}
\label{atom_opt}
& \underset{d_j \in \R^p}{\text{min}}
&& \|F - d_j \, x_{j,{\cal I}_j}\|_F^2
\end{aligned}
\ee
The norm constraint $\|d_j\|_2 = 1$ is usually imposed after solving the optimization problem.

{\em AK-SVD.}
The K-SVD algorithm and its approximate version AK-SVD \cite{RZE08} treat (\ref{atom_opt})
by considering that $x_{j,{\cal I}_j}$ is also a variable.
Problem (\ref{atom_opt}) becomes a rank-1 approximation problem that is solved
by AK-SVD with a single iteration of the power method (to avoid ambiguity, we add
superscripts representing the iteration number):
\be
\begin{aligned}
& d_j^{(k+1)} = F (x_{j,{\cal I}_j}^{(k)})^T / \| F (x_{j,{\cal I}_j}^{(k)})^T \|_2 \\
& x_{j,{\cal I}_j}^{(k+1)} = F^T d_j^{(k+1)}
\end{aligned}
\label{eq:pak-svd-solution}
\ee
Note that the representations are also changed in the atom update stage,
which is the specific of this approach.

\smallskip
{\em SGK.}
Dictionary learning for sparse representations as a generalization of K-Means
clustering (SGK) \cite{SM13} solves directly problem (\ref{atom_opt}).
This is a least squares problem whose solution is
\be
d_j^{(k+1)} = F x_{j,{\cal I}_j}^T / (x_{j,{\cal I}_j} x_{j,{\cal I}_j}^T).
\label{eq:sgk-solution}
\ee

The atom updates part of the general JAU scheme from algorithm \ref{algo:dl-jau}
has the form described by algorithm \ref{algo:pa-sgk}, named P-SGK (with P from Parallel).
The error $E$ is recomputed in step 2 before each group of atom updates,
thus taking into account the updated values of the previous groups.
Depending on the value of $\tilde n$, the error can be computed more efficiently
via updates to its previous value instead of a full recomputation.
Steps 4 and 5 implement relations (\ref{Ferror}) and (\ref{eq:sgk-solution}), respectively.
Step 6, the normalization, is identical with that from the general scheme.

To obtain the JAU version of AK-SVD (named PAK-SVD), we replace step 4
by the operations from (\ref{eq:pak-svd-solution}).
Note that, for full parallelism ($\tilde n = n$),
P-SGK and PAK-SVD are identical, since the atoms produced by
(\ref{eq:pak-svd-solution}) and (\ref{eq:sgk-solution}) have the same direction.
For full parallelism the representations updated by PAK-SVD are not used,
while if $\tilde n < n$, some updated representations affect the error matrix
from step 2.

\begin{algorithm}[t]
\label{algo:pa-sgk}
\SetKwComment{Comment}{}{}

\BlankLine
\KwData{\quad current dictionary $D \in \R^{p \times n}$ \\
	\hspace{13.6mm} signals set $Y \in \R^{p \times m}$ \\
	\hspace{13.6mm} sparse representations $X \in \R^{n \times m}$ \\
	\hspace{13.6mm} number of parallel atoms $\tilde n$}
\KwResult{\hspace{1mm}next dictionary $D$}
\BlankLine

\For{$\ell = 1$ \KwTo $n / \tilde n$}{
	$E = Y - D X$ \\
	\For{$j = (\ell-1) \tilde n + 1$ \KwTo $\ell \tilde n$, in parallel}{
		$F = E_{{\cal I}_j} + d_j x_{j,{\cal I}_j}$ \\
		$d_j = F x_{j,{\cal I}_j}^T / (x_{j,{\cal I}_j} x_{j,{\cal I}_j}^T)$ \\
		$d_j \leftarrow d_j/\|d_j\|_2$ \\
	}
}
\caption{P-SGK Atom Updates}
\end{algorithm}


\smallskip
{\em NSGK.}
The update problem (\ref{atom_opt}) is treated in \cite{SBJ13} in terms of
differences with respect to the current dictionary and representations,
instead of working directly with $D$ and $X$.
Applying this idea to SGK, the optimization problem is similar, but with the
signal matrix $Y$ replaced by
\be
Z = Y + D^{(k)} X^{(k-1)} - D^{(k)} X^{(k)}
\label{Zerr}
\ee
where $X^{(k-1)}$ is the sparse representation matrix at the beginning of
the $k$-iteration of the DL algorithm, while $X^{(k)}$ is the matrix
computed in the $k$-th iteration
(e.g.\ in step 1 of Algorithm~\ref{algo:dl-jau}).
The P-NSGK algorithm (NSGK stands for New SGK, the name used in \cite{SBJ13})
is thus identical with P-SGK, with step 2 modified according to (\ref{Zerr}).
Also, in (\ref{eq:sgk-solution}), the representations $x_{j,{\cal I}_j}$
are taken from $X^{(k-1)}$, not from $X^{(k)}$ as for the other methods.

\section{Numerical results}
\label{sec:res}

We give here numerical evidence supporting the advantages of the JAU scheme,
for two standard problems: recovery of a given dictionary and DL for sparse
image representation.
We compare the JAU algorithms PAK-SVD, P-SGK and P-NSGK
with their sequential counterparts.
We report results obtained with the same input data for all the algorithms;
in particular, the initial dictionary is the same.
The sparse representations were computed via OMP\footnote{
We used OMP-Box version 10 available at
\url{http://www.cs.technion.ac.il/~ronrubin/software.html}}.

\subsection{Dictionary recovery}

Following the numerical experiments from \cite{SM13} and \cite{SBJ13},
we generated a random dictionary with $n=50$ atoms of size $p=20$ each,
and a signal set $Y$ of $m=1500$ data vectors, each vector being generated as a
linear combination of $s \in \{3,4,5\}$ different atoms.
We then added white gaussian noise with SNR levels of
$10$, $20$, $30$ and $\infty$ dB to the signal set.
We applied $9s^2$ dictionary learning iterations on this signal set for each
algorithm and compared the resulting dictionaries with the original
in the same way as in \cite{AEB06}.
The dictionary was initialized with a random selection of data vectors.
The algorithms were given the fixed sparsity target $s$
that was used to generate the original signal set.
JAU methods used full parallelism ($\tilde n = n$).
The percentages of recovered atoms, averaged over 50 runs, are presented in table \ref{tab:recov}.
(PAK-SVD is not reported, since it gives the same results as P-SGK.)

We note that, although the JAU algorithms give the best result in 6 out of the 12
considered problems, the results are rather similar for all algorithms.
(We may infer that, in this problem, the sparse representation stage is the bottleneck,
not the atom update stage.)
We can at least conclude that, for dictionary recovery, the JAU scheme is
not worse than the sequential ones.

\begin{table}
	\tabcolsep 5pt
	\caption{Percentage of recovered atoms}
    \label{tab:recov}
		\small
	\bc \bt{ |c |  c | c | c | c | c | }
		\hline
		$s$ & Method & \multicolumn{4}{c|}{$SNR$} \\
		& & \multicolumn{1}{c}{10} & \multicolumn{1}{c}{20}
		& \multicolumn{1}{c}{30} & \multicolumn{1}{c|}{$\infty$} \\
		\hline

		\multirow{5}{*}{$3$}
		& NSGK		& 87.16 & {\bf 90.16} & 89.32 & 89.56 \\
		& P-NSGK	& {\bf 88.36} & 89.64 & 89.92 & 89.76 \\
		& SGK		& 87.44 & 89.40 & 88.80 & {\bf 90.12} \\
		& P-SGK		& 86.48 & 89.84 & 89.00 & 88.24 \\
		& AK-SVD	& 86.20 & 90.00 & {\bf 90.00} & 88.52 \\
		\hline

		\multirow{5}{*}{$4$}
		& NSGK		& {\bf 70.68} & 91.88 & 92.16 & 93.16 \\
		& P-NSGK   	    & 68.08 & 92.48 & {\bf 92.88} & {\bf 93.28} \\
		& SGK		    & 67.28 & 92.16 & 91.68 & 91.92 \\
		& P-SGK		& 68.28 & {\bf 93.48} & 91.88 & 92.56 \\
		& AK-SVD	& 70.08 & 92.76 & 92.16 & 92.32 \\
		\hline

		\multirow{5}{*}{$5$}
		& NSGK		& 10.24 & 92.36 & 92.72 & 94.40 \\
		& P-NSGK	& 10.60 & 93.08 & {\bf 93.32} & 94.72 \\
		& SGK		& 11.68 & {\bf 93.28} & 92.60 & 93.92 \\
		& P-SGK		& {\bf 12.04} & 93.00 & 92.92 & 93.92 \\
		& AK-SVD	& 11.64 & 92.72 & 92.88 & {\bf 94.96} \\
		\hline
	\et \ec
\end{table}

\subsection{Dictionary learning}

The training signals were images taken from the USC-SIPI \cite{sipi} database
(e.g.\ barb, lena, boat, etc.).
The images were normalized and split into random $8 \times 8$ blocks.
The initial dictionary was built with random atoms.

\begin{figure}
\bc \framebox{\includegraphics[width=0.90\linewidth]{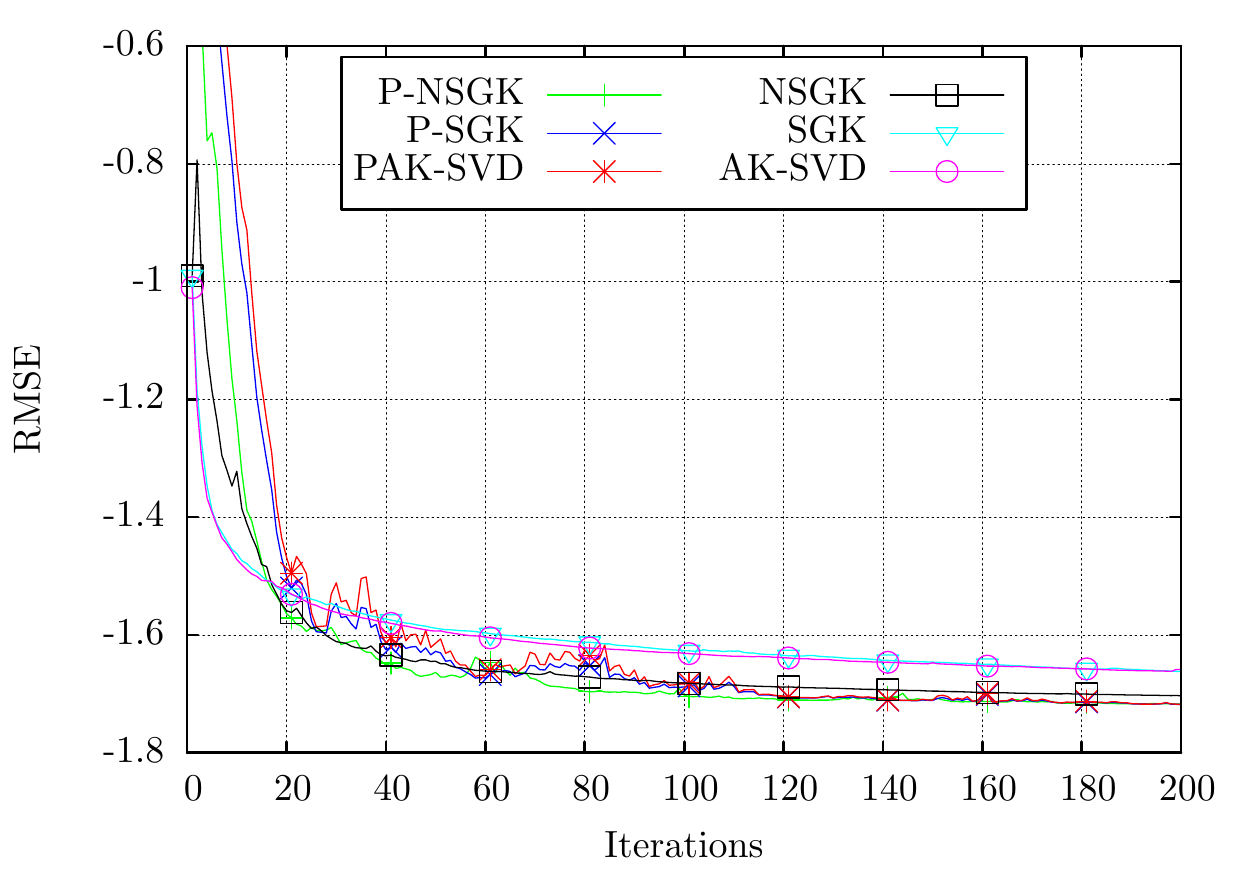}} \ec
\caption{Error evolution for parallel and sequential algorithms.}
\label{fig:pa-rmse}
\end{figure}

In a first experiment, we used $m=32768$ signals of dimension $p=64$ to train
dictionaries with $n=512$ atoms, with a target sparsity of $s=8$.
In figure \ref{fig:pa-rmse} we can see the evolution of the representation RMSE,
averaged over 10 runs, for the JAU and sequential algorithms.
JAU algorithms have full parallelism ($\tilde n = n$).
(Note that the PAK-SVD and P-SGK curves are slightly different, due to the computation
of the errors at the end of a DL iteration, where PAK-SVD has different representations;
otherwise, the dictionaries are identical.)
Although the sequential algorithms have smoother convergence, the proposed
parallel versions obtain clearly better results.
Among the sequential algorithms, NSGK is the best, confirming the findings from \cite{SBJ13}.
However, all parallel algorithms are better than NSGK.

The same conclusion is supported by a second experiment, where the conditions are
similar but, for faster execution, only $m=16384$ training signals were used.
Table \ref{tab:rmse} shows the lowest RMSE after $k=200$ iterations, averaged over 10 runs,
for three values of the dictionary size $n$.
In all cases, the JAU algorithms are clearly the best.

\begin{table}
	\tabcolsep 5pt
	\caption{Best RMSE values after 200 iterations}
	\label{tab:rmse}
		\small
	\bc \bt{ | c |  c | c | c | }
		\hline
		& $n=128$ & $n=256$ & $n=512$ \\
		\hline
		NSGK		& 0.0185 & 0.0168 & 0.0154 \\
		\hline
		P-NSGK		& 0.0167 & 0.0154 & 0.0139 \\
		\hline
		SGK		& 0.0201 & 0.0185 & 0.0166 \\
		\hline
		P-SGK		& 0.0165 & 0.0153 & 0.0138 \\
		\hline
		AK-SVD		& 0.0201 & 0.0184 & 0.0163 \\
		\hline
	\et \ec
\end{table}

\begin{figure}
\bc \framebox{\includegraphics[width=0.90\linewidth]{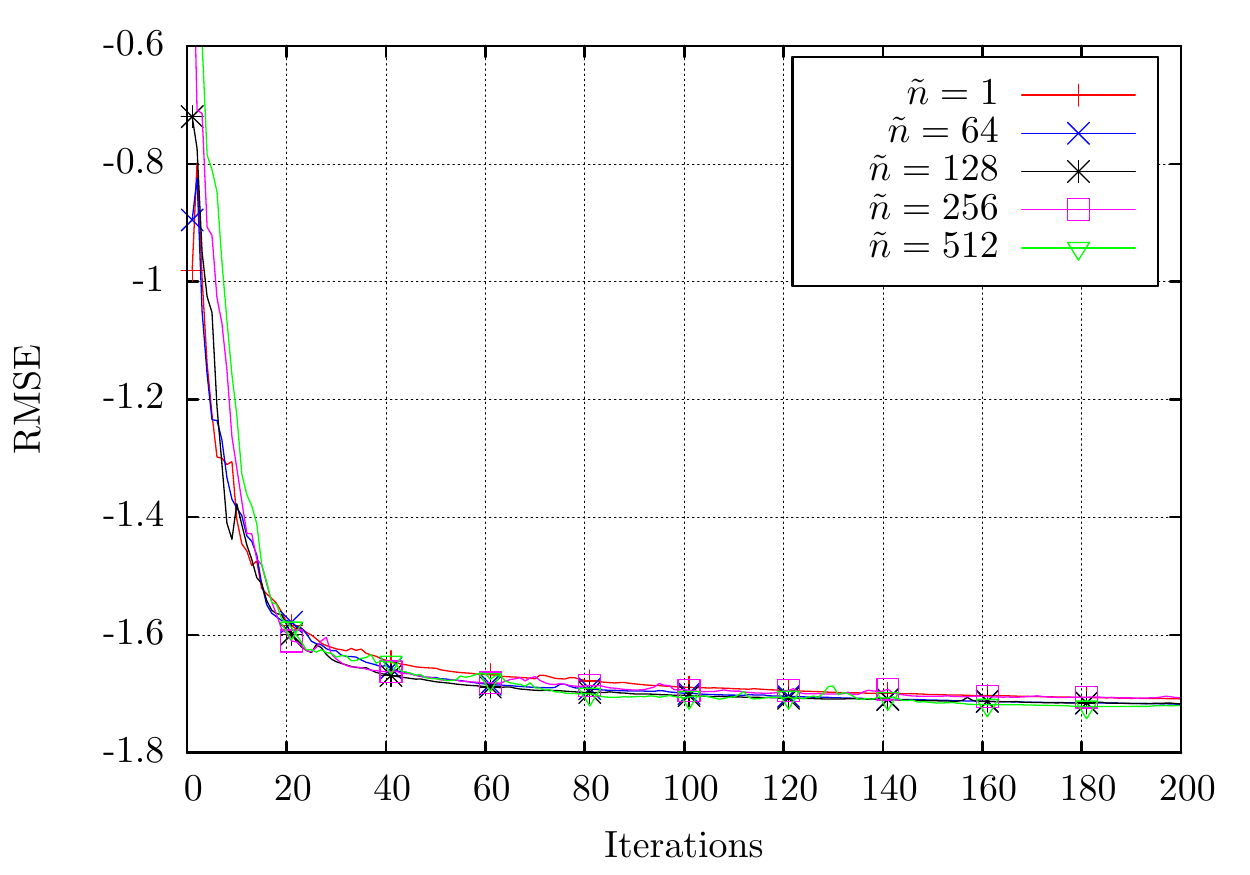}} \ec
\caption{P-NSGK error evolution for various group sizes.}
\label{fig:pa-nsgk}
\end{figure}

To show the influence of the group size $\tilde n$ of the JAU algorithms,
we present
the evolution of the error, averaged over 10 runs, in figure \ref{fig:pa-nsgk}.
We see that the effect of group size is less intuitive.
Full parallelism ($\tilde n = n$) is the winner, although some smaller
values of $\tilde n$ are good competitors, almost all being better
than the sequential version ($\tilde n = 1$).
Although in this example $\tilde n = 256$ is worse than some smaller values,
the error usually decreases as $\tilde n$ grows, the best
value being $\tilde n = n$ in all our tests, for all parallel methods.
A possible explanation is that the JAU strategy, due to the independent
atom updates, is less prone to get trapped in local minima.
Modifying atoms one by one, although locally optimal, may imply only small
modifications of the atoms; in contrast, JAU appears to be able of
larger updates that make convergence more erratic, but can reach
a better dictionary.

\subsection{JAU versus MOD}

We now compare the performance in representation error of
the JAU algorithms with the intrinsically parallel algorithm
named method of optimal directions (MOD) \cite{EAH99mod}.
MOD uses OMP for representation and updates
the dictionary $D$ with the least-squares solution of the linear
system $DX = Y$.
For completeness we also include the sequential versions
on which JAU algorithms are built.

\begin{figure}
\bc \framebox{\includegraphics[width=0.90\linewidth]{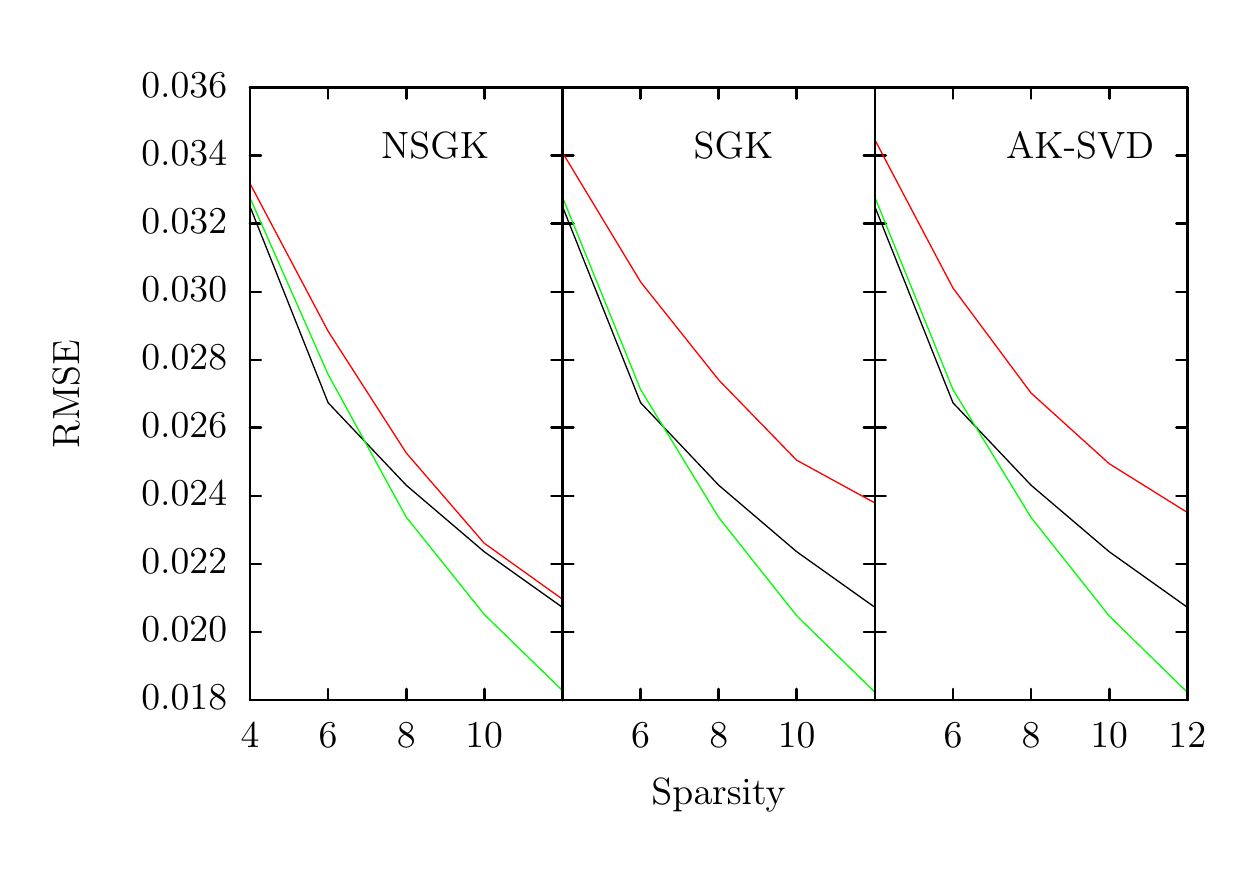}} \ec
\caption{Final errors for different sparsity constraints.
The sequential versions are red, JAU algorithms are green and MOD is black.}
\label{fig:avg-s}
\end{figure}

\begin{figure}
\bc \framebox{\includegraphics[width=0.90\linewidth]{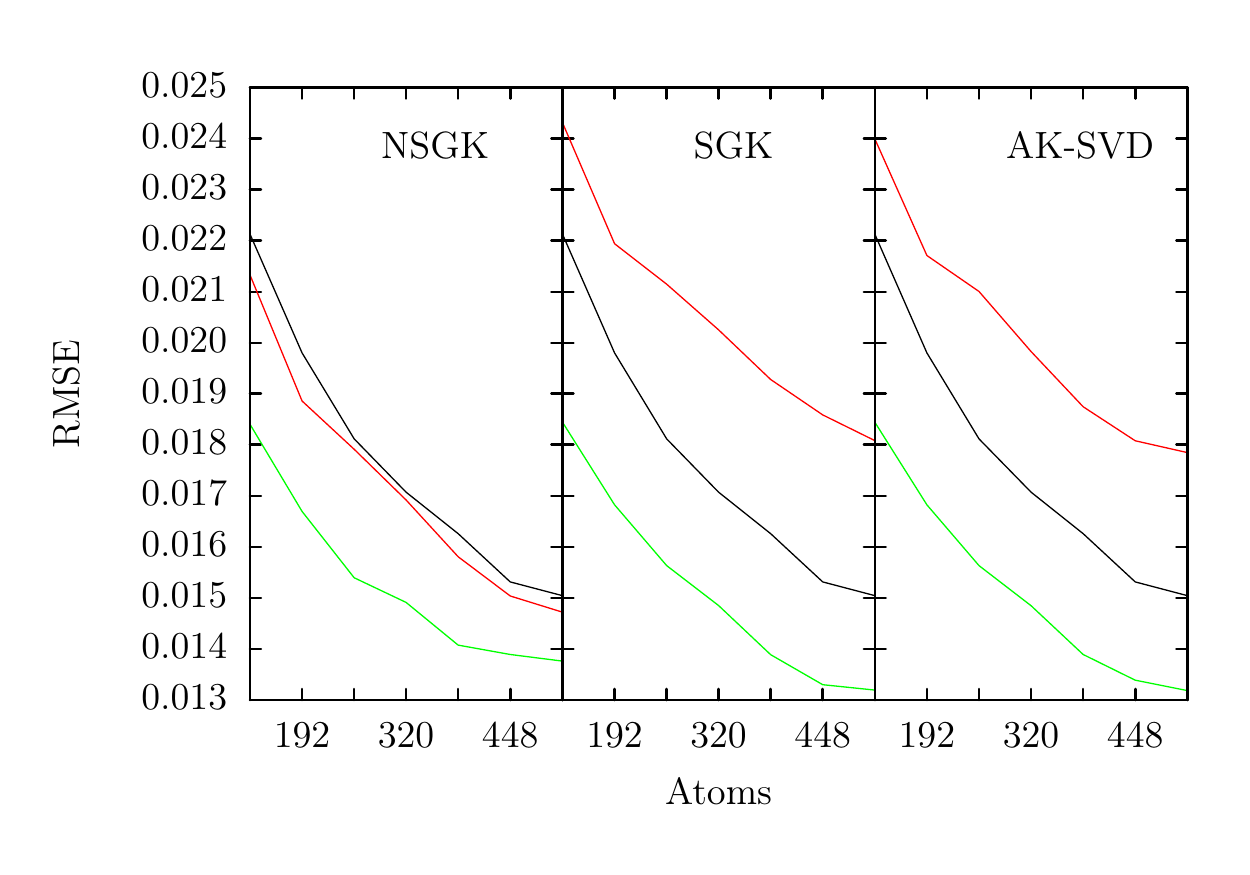}} \ec
\caption{Final errors for varied dictionary sizes.
The sequential versions are red, JAU algorithms are green and MOD is black.}
\label{fig:avg-n}
\end{figure}

\begin{figure}
\bc \framebox{\includegraphics[width=0.90\linewidth]{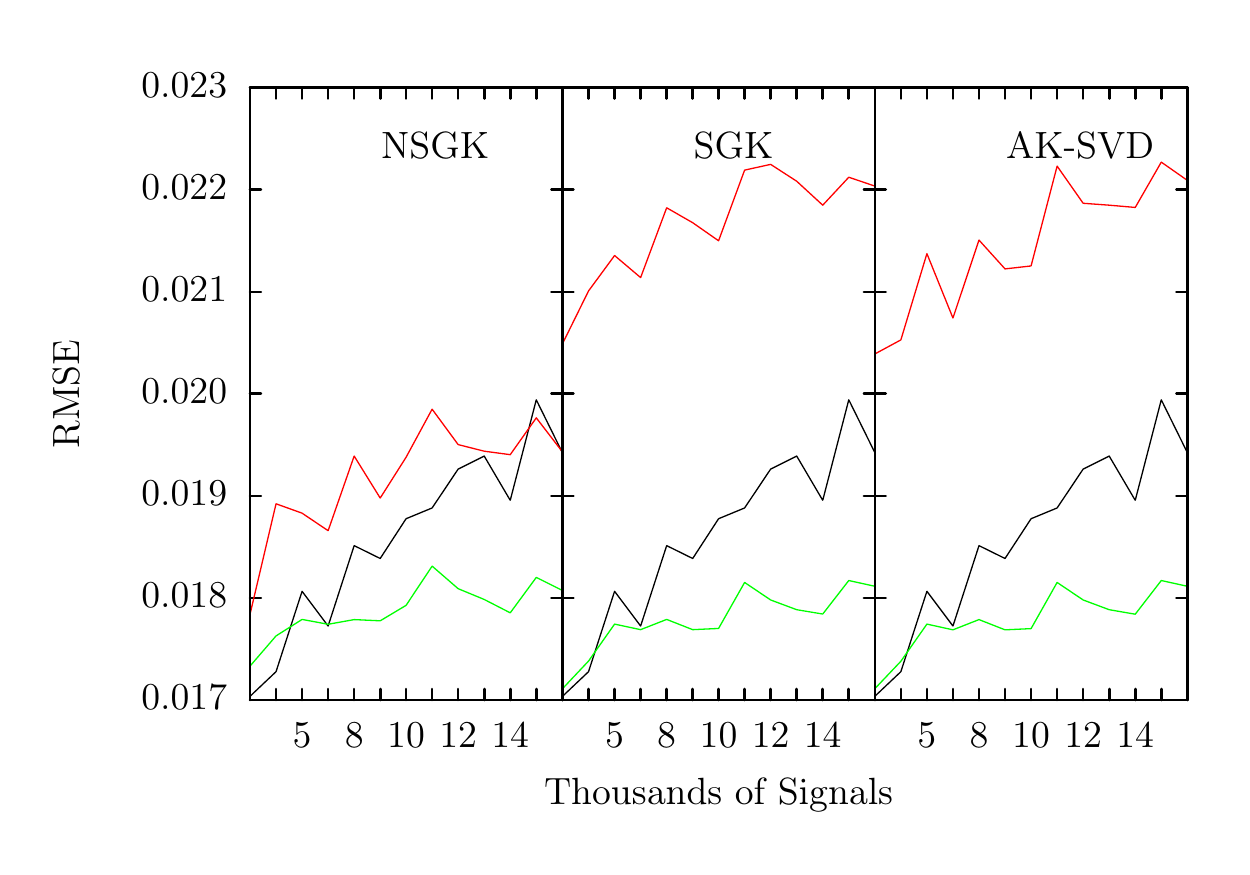}} \ec
\caption{Final errors for varied training set sizes.
The sequential versions are red, JAU algorithms green and MOD is black.}
\label{fig:avg-m}
\end{figure}

\begin{figure}
\bc \framebox{\includegraphics[width=0.90\linewidth]{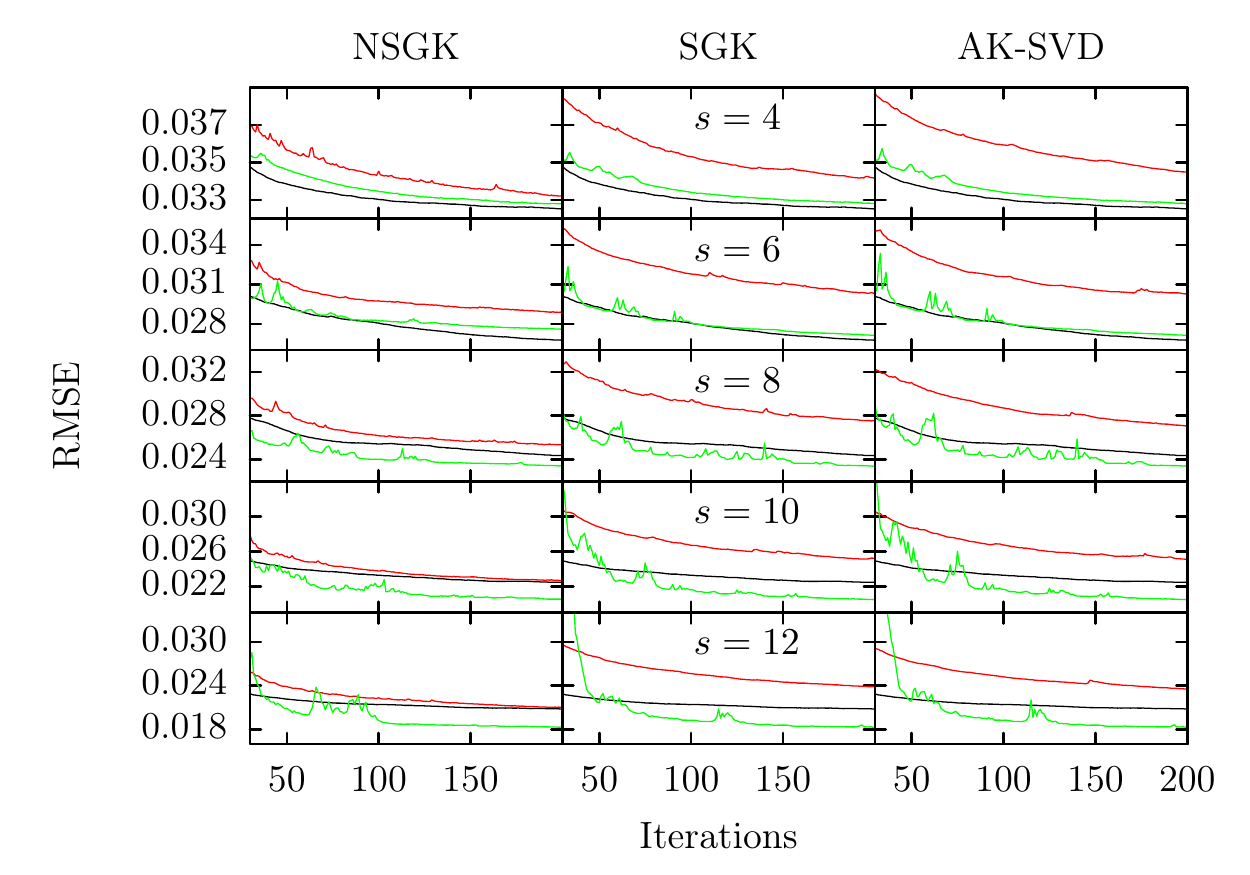}} \ec
\caption{Error evolution at different sparsity constraints.
The sequential versions are red, JAU algorithms green and MOD is black.}
\label{fig:error-evol}
\end{figure}

In figures \ref{fig:avg-s}--\ref{fig:error-evol}
we depict the JAU algorithms with green,
the sequential versions with red
and MOD with black.
All algorithms performed DL for $k=200$ iterations.
Each data point from these figures represents an average of 10 runs
of the same algorithm with the same parametrization and data dimensions
but with training sets composed of different image patches.

To see how sparsity influences the end result,
figure \ref{fig:avg-s} presents the final errors for
sparsity levels starting from $s=4$ up to $s=12$
when performing DL for dictionaries of $n=128$ atoms
on training sets of size $m=8192$.
We notice that for all three algorithms (NSGK, SGK and AK-SVD)
the JAU methods perform similar to MOD at lower sparsity constraints,
but as we pass $s=8$ our proposed parallel strategy is clearly better.
The sequential versions always come in last,
except perhaps for NSGK that comes close to MOD past $s=10$.

Figure \ref{fig:avg-n} presents the final errors
for DL on training sets of $m=12288$ signals,
with a sparsity constraint of $s=12$,
when varying the total number of atoms in the dictionary
from $n=128$ to $n=512$ in increments of 64.
Again, the JAU versions are the winners for all three algorithms.
Out of the sequential algorithms,
NSGK is the only one that manages to out-perform MOD,
while the others lag behind coming in last.

The next experiment investigates the influence of the signal set size
on the final errors.
In figure \ref{fig:avg-m} we kept
a fixed dictionary size of $n=256$ and a sparsity of $s=10$
and performed DL
starting with training sets of $m=4096$ signals
that we increased in increments of 1024 up to $m=16384$.
JAU stays ahead of MOD almost everywhere, except for small signal sets
with $m<5000$ where the results are similar.
The sequential versions are once again the poorest performers.

Finally, we present in figure \ref{fig:error-evol}
the error improvement at each iteration for all algorithms,
for several sparsity levels.
In this experiment
we used a dictionary of $n=128$ atoms
and a training set of $m=8192$ signals.
We can see that the JAU versions can jump back and forwards,
specially during the first iterations.
We think that this is due to the parallel update of the dictionary atoms
which leads to jumps from one local minima to another
until a stable point is reached.
This is, perhaps, the reason why in the end it manages to provide a lower
representation error.
Even though the JAU convergence is not as smooth as MOD or the sequential
versions, it has a consistent descendent trend.

\subsection{Execution times}

We now focus on the improvements in execution time.
We used OpenCL for our GPU implementation of the JAU algorithms
and performed the execution on an
ATI FirePro V8800 (FireGL V) card from AMD,
running at a maximum clock frequency of 825MHz,
having 1600 streaming processors,
2GB global memory and 32KB local memory.
For the sequential versions we used a C implementation that
kept identical instructions everywhere where possible
in order to provide an accurate comparison
that clearly shows the improvements in execution time brought,
almost exclusively, by the JAU strategy.
The sequential tests were executed on an Intel i7-3930K CPU
running at a maximum clock frequency of 3.2GHz.

\begin{figure}[t]
\bc \framebox{\includegraphics[width=0.90\linewidth]{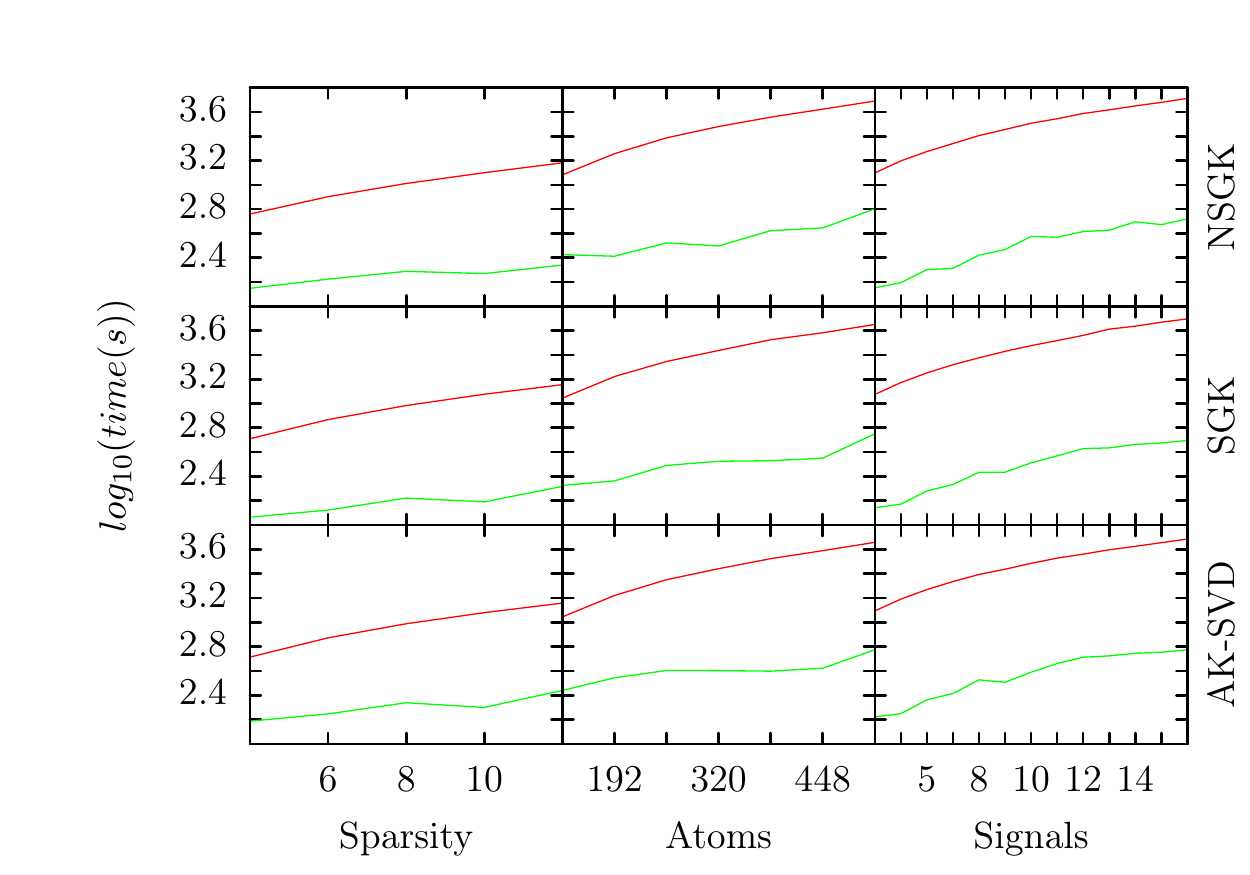}} \ec
\caption{Execution times.
The sequential versions are red and the JAU algorithms are green.}
\label{fig:perf}
\end{figure}

We present 3 experiments in figure \ref{fig:perf}
where we vary
the sparsity constraint,
the atoms in the dictionary
and the number of signals in the training set.
We depict the JAU versions with green and the sequential versions with red.
Because of the significant difference in execution time,
we use a logarithmic scale.
Again, we used $k=200$ iterations for all methods.

For the sparsity experiment we used
a dictionary of $n=128$ and a training set of $m=8192$
and we increased the sparsity from $s=4$ to $s=12$ in increments of 2.
When studying the dictionary impact on the execution performance we kept
a fixed training set of $m=12288$
and a sparsity of $s=6$
and varied the atoms from $n=128$ in increments of 64 up to $n=512$.
Finally, we increased the signal set in increments of 1024 starting
from $m=4096$ until $m=16384$
with a fixed dictionary of $n=256$ and a sparsity of $s=10$.
In the panel the abscissa tics represent thousands of signals.

In all of our experiments the JAU versions showed important
improvements in execution time, the speed-up reaching values as high as
10.6 times for NSGK,
10.8 times for SGK and
12 times for AK-SVD.
This was to be expected, since JAU algorithms are naturally parallel
in the atom update stage.

\section{Conclusions}

We have shown that several dictionary learning algorithms, like
AK-SVD \cite{RZE08}, SGK \cite{SM13} and NSGK \cite{SBJ13},
benefit from adopting Jacobi (parallel) atom updates instead of
the usual Gauss-Seidel (sequential) ones.
We have also shown that the new Jacobi algorithms outperform
their sequential standard versions and also other types of algorithms
like MOD \cite{EAH99mod}.
In the mostly academic dictionary recovery problem, the
parallel and sequential versions have similar performance.
However, in the more practical problem of dictionary learning
for sparse image representation, the proposed parallel algorithms
have a clearly better behavior with superior execution times.

\bibliography{sparse_approx}

\begin{thebibliography}{10}
\expandafter\ifx\csname url\endcsname\relax
  \def\url#1{\texttt{#1}}\fi
\expandafter\ifx\csname urlprefix\endcsname\relax\def\urlprefix{URL }\fi
\expandafter\ifx\csname href\endcsname\relax
  \def\href#1#2{#2} \def\path#1{#1}\fi

\bibitem{AEB06}
M.~Aharon, M.~Elad, A.~Bruckstein, {K-SVD: An Algorithm for Designing
  Overcomplete Dictionaries for Sparse Representation}, IEEE Trans. Signal
  Proc. 54~(11) (2006) 4311--4322.

\bibitem{RZE08}
R.~Rubinstein, M.~Zibulevsky, M.~Elad, {Efficient Implementation of the K-SVD
  Algorithm using Batch Orthogonal Matching Pursuit}, Tech. Rep. CS-2008-08,
  Technion Univ., Haifa, Israel (2008).

\bibitem{EAH99mod}
K.~Engan, S.~Aase, J.~Husoy, {Method of optimal directions for frame design},
  in: IEEE Int. Conf. Acoustics Speech Signal Proc., Vol.~5, 1999, pp.
  2443--2446.

\bibitem{PRK93omp}
Y.~Pati, R.~Rezaiifar, P.~Krishnaprasad, {Orthogonal matching pursuit:
  Recursive function approximation with applications to wavelet decomposition},
  in: 27th Asilomar Conf. Signals Systems Computers, Vol.~1, 1993, pp. 40--44.

\bibitem{SmEl13}
L.~Smith, M.~Elad, {Improving Dictionary Learning: Multiple Dictionary Updates
  and Coefficient Reuse}, IEEE Signal Proc. Letters 20~(1) (2013) 79--82.

\bibitem{SM13}
S.~K. Sahoo, A.~Makur, {Dictionary training for sparse representation as
  generalization of $K$-Means clustering}, Signal Processing Letters, IEEE
  20~(6) (2013) 587--590.

\bibitem{SBJ13}
M.~Sadeghi, M.~Babaie-Zadeh, C.~Jutten, {Dictionary Learning for Sparse
  Representation: a Novel Approach}, IEEE Signal Proc. Letter 20~(12) (2013)
  1195--1198.

\bibitem{SBJ14}
M.~Sadeghi, M.~Babaie-Zadeh, C.~Jutten, {Learning Overcomplete Dictionaries
  Based on Atom-by-Atom Updating}, IEEE Trans. Signal Proc. 62~(4) (2014)
  883--891.

\bibitem{RuDu12spl}
C.~Rusu, B.~Dumitrescu, {Stagewise K-SVD to Design Efficient Dictionaries for
  Sparse Representations}, IEEE Signal Proc. Letters 19~(10) (2012) 631--634.

\bibitem{SDB12}
C.~Sigg, T.~Dikk, J.~Buhmann, {Learning Dictionaries With Bounded
  Self-Coherence}, IEEE Signal Proc. Letters 19~(19) (2012) 861--865.

\bibitem{BaPl13}
D.~Barchiesi, M.~Plumbley, {Learning Incoherent Dictionaries for Sparse
  Approximation Using Iterative Projections and Rotations}, IEEE Trans. Signal
  Proc. 61~(8) (2013) 2055--2065.

\bibitem{RBE10}
R.~Rubinstein, A.~Bruckstein, M.~Elad, {Dictionaries for Sparse Representations
  Modeling}, Proc. IEEE 98~(6) (2010) 1045--1057.

\bibitem{ToFr11}
I.~Tosic, P.~Frossard, {Dictionary Learning}, IEEE Signal Proc. Mag. 28~(2)
  (2011) 27--38.

\bibitem{IrDu14}
P.~Irofti, B.~Dumitrescu, {GPU Parallel Implementation of the Approximate K-SVD
  Algorithm Using OpenCL}, in: EUSIPCO, Lisbon, Portugal, 2014.

\bibitem{sipi}
A.~Weber, \href{http://sipi.usc.edu/database}{{The USC-SIPI Image Database}}
  (1997).
\newline\urlprefix\url{http://sipi.usc.edu/database}

\end{thebibliography}

\end{document}